\documentclass[runningheads]{llncs}
\usepackage{color}
\usepackage{graphicx}
\usepackage[misc,geometry]{ifsym}
\begin{document}
\title{Which Parts Determine\\ the Impression of the Font?}
\titlerunning{Which Parts Determine the Impression of the Font?}
%\vspace{-1cm}

\author{Masaya Ueda\inst{1}(\Letter) \and
Akisato Kimura\inst{2} \and
Seiichi Uchida\inst{1}\orcidID{0000-0001-8592-7566}}

\authorrunning{M. Ueda et al.}
\institute{Kyushu University, Fukuoka, Japan
\email{\{masaya.ueda\}@human.ait.kyushu-u.ac.jp}\and
NTT Communication Science Laboratories, NTT Corporation, Japan}
\maketitle              % typeset the header of the contribution
%
%\vspace{-1cm}
\begin{abstract}
%The abstract should briefly summarize the contents of the paper in 15--250 words.
Various fonts give different impressions, such as legible, rough, and comic-text. This paper aims to analyze the correlation between the local shapes, or parts, and the impression of fonts. By focusing on local shapes instead of the whole letter shape, we can realize more general analysis independent from letter shapes. The analysis is performed by newly combining SIFT and DeepSets, to extract an arbitrary number of essential parts from a particular font and aggregate them to infer the font impressions by nonlinear regression. Our qualitative and quantitative analyses prove that (1)~fonts with similar parts have similar impressions, (2)~many impressions, such as legible and rough, largely depend on specific parts, and (3)~several impressions are very irrelevant to parts.
\end{abstract}
%
%
%%%%%%%%%%%%%%%%%%%%%%%%%%%%%%%%%%%%%%%%%%%%%%%%%%%%%%%%%%%
\section{Introduction} \label{sec:intro}
\begin{figure}[t]
    \centering
%    \vskip 5cm
    \includegraphics[width=1.0\linewidth]{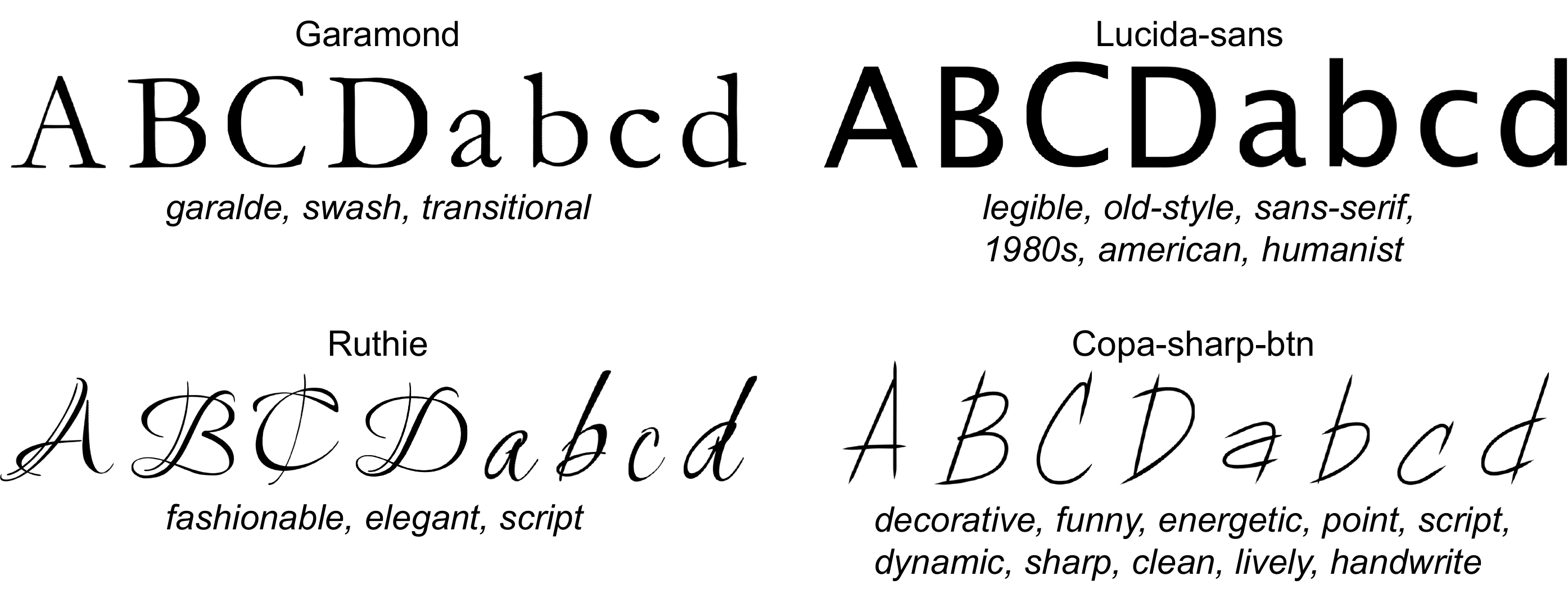}\\[-5mm]
    \caption{Fonts and their impressions (from \cite{Chen2019large}).}
%    \vspace{-5mm}
    \label{fig:shape-and-impression}
\end{figure}

%%%%%%%%%%%%%%%%%%%%%%%%%%%%%%%%%%%%%%%%%%%%%%%%%%%%%%%%%%%
Different font shapes will give different impressions. Fig.~\ref{fig:shape-and-impression} shows several font examples and their impressions attached by human annotators~\cite{Chen2019large}. A font ({\tt Garamond}) gives a {\it traditional} impression and another font ({\tt Ruthie}) an {\it elegant} impression. Fig.~\ref{fig:shape-and-impression} also shows that multiple impressions are given to a single font. Note that the meaning of the term ``impression'' is broader than usual in this paper; it refers to not only (more subjective) actual impressions, such as {\it elegant}, but also (less subjective) words of font shape description, such as {\it sans-serif}. 
\par
%---------------------------------
%---------------------------------
The relationship between fonts and their impressions is not well explained yet, 
despite many attempts from 100 years ago (e.g., \cite{poffenberger1923study} in 1923).
This is because the past attempts were subjective and small-scale. Moreover, 
we need to deal with the strongly nonlinear relationship between fonts and impressions and  find image features that are useful for explaining the relationship.\par
Fortunately, recent image analysis and machine learning techniques provide a reliable and objective analysis of complex nonlinear relationships. In addition, a  large font image dataset with impression annotations is available now by Chen et al.~\cite{Chen2019large}; their dataset contains 18,815 fonts, and several impression words (from 1,824 vocabularies) are attached to each font. The examples of Fig.~\ref{fig:shape-and-impression} are taken from this dataset.\par
The purpose of this paper is to analyze the relationship between fonts and their impressions as objectively as possible by using the large font image dataset~\cite{Chen2019large} with a machine learning-based approach. The relationship revealed by our analysis will help design a new font with a specific impression and judge the font appropriateness in a specific situation. Moreover, it will give hints to understand the psychological correlation between shape and impression.\par
We focus on local shapes, or {\em parts}, formed by character strokes for the relationship analysis. Typical examples of parts are the endpoints, corners, curves, loops (i.e., holes) and intersections. These parts will explain the relationship more clearly and appropriately than the whole character shape because of the following three reasons. First, the whole character shape is strongly affected by the character class, such as `A' and `Z,' whereas the parts are far less. Second, the decorations and stylizations are often attached to parts, such as serif, rounded corners, and uneven stroke thickness. Third, part-based analysis is expected to provide more explainability because it can localize the part that causes the impression. Although it is also true that the parts cannot represent some font properties, such as the whole character width and the aspect ratio, they are still suitable for analyzing the relationship between shape and impression, as proved by this paper.\par
Computer vision research in the early 2000s often uses parts of an image, called {\em keypoints}, as features for generic object recognition. Using some operators, such as the Difference of Gaussian (DoG) operator, the $L$ informative parts (e.g., corners) are first detected in the input image. Then, each detected part is described as a feature vector that represents the local shape around the part. Finally, the image is represented as a set of $L$ feature vectors. As we will see in Section~\ref{sec:related}, many methods, such as SIFT~\cite{SIFT} and SURF~\cite{SURF}, have been proposed to detect and describe the parts.  \par
This paper proposes a new approach where this well-known local descriptor is integrated into a recent deep learning-based framework, called DeepSets~\cite{deepsets}.
DeepSets can accept an arbitrary number of $D$-dimensional input vectors. Let $\mathbf{x}_1,\ldots,\mathbf{x}_L$ denote the input vectors (where $L$ is variable). In DeepSets, the input vectors are converted into another vector representation,   $\mathbf{y}_1,\ldots,\mathbf{y}_L$ by a certain neural network $g$, that is, $\mathbf{y}_l=g(\mathbf{x}_l)$. Then, they are {\em summarized} as a single vector by taking their sum\footnote{As noted in \cite{deepsets}, it is also possible to use another operation than the summation, such as element-wise max operation.}, i.e., $\tilde{\mathbf{y}}=\sum_l \mathbf{y}_l$. Finally, the vector $\tilde{\mathbf{y}}$ is fed to another neural network $f$ that gives the final output $f(\tilde{\mathbf{y}})$.\par
\begin{figure}[t]
    \centering
    \includegraphics[width=1\linewidth]{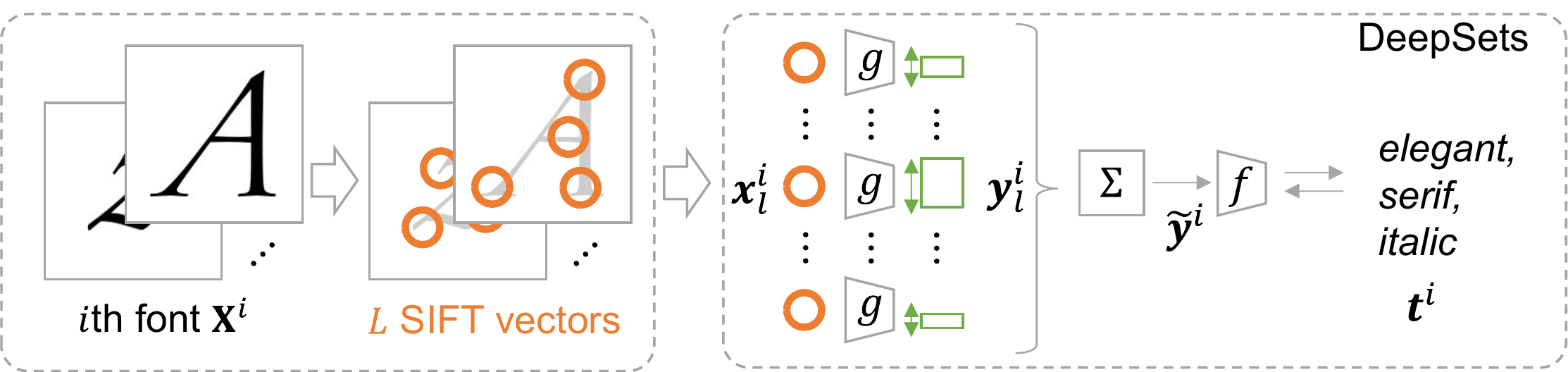}\\[-3mm]
    \caption{Overview of the part-based impression estimation by DeepSets.}
    \label{fig:overall}
    %\vspace{-5mm}
\end{figure}
As shown in Fig~\ref{fig:overall}, we use DeepSets to estimate a font's impressions by a set of its parts. Precisely, each input vector $\mathbf{x}^i_l$ of DeepSets corresponds to the $D$-dimensional SIFT descriptor representing a part detected in the $i$th font. The output $f(\tilde{\mathbf{y}}^i)$ is the vector showing the $K$-dimensional impression vector, where $K$ is the vocabulary size of the impression words. The entire network of DeepSets, i.e., $f$ and $g$, is trained to output the $K$-dimensional $m^i$-hot vector for the font with $m^i$ impression words in an end-to-end manner.\par
The above simple framework of DeepSets is very suitable for our relation analysis task due to the following two properties. First, DeepSets is invariant to the order of the $L$ input vectors. Since there is no essential order of the parts detected in a font image, this property allows us to feed the input vectors to DeepSets without any special consideration. \par
%==== ここでの the relation analysis task は何を指すものでしょうか？ 同じ言葉が使われた task がすぐには見当たらないので．．． Kimura 11:22, Feb 18, 2021.
%
The second and more important property of DeepSets is that it can learn the {\em importance of the individual parts} in the impression estimation task. If a part $\mathbf{x}^i_l$ is not important for giving an appropriate estimation result, its effect will be weakened by minimizing the norm of $\mathbf{y}^i_l$. Ultimately, if $\|\mathbf{y}^i_l\|$ becomes zero by the representation network $g$, the part $\mathbf{x}^i_l$ is totally ignored in the estimation process. 
Therefore, we can understand which part is important for a specific impression by observing the norm $\|\mathbf{y}^i_l\|$.
This will give a far more explicit explanation of the relationship between shape and impression than, for example, the style feature that is extracted from the whole character shape by disentanglement (e.g.,~\cite{Liu2018}).  
\par
The main contributions of this paper are summarized as follows:
\begin{itemize}
    \item This paper proves that a specific impression of a font largely correlates to its local shapes, such as corners and endpoints of strokes. This is the first proof of this correlation by an objective and statistical analysis with a large-scale dataset to the authors' best knowledge.
    \item  The well-known local descriptor called SIFT is integrated into a recent deep learning-based framework called DeepSets for the part-based impression analysis. DeepSets has appropriate properties for the task; especially, it allows us to evaluate the importance of each part for a specific impression.  
    \item The analysis results provide various findings of the correlation between local shapes and impressions. For example, we could give an answer to the well-known open-problem; {\it what is the legibility?} Our analysis results show that constant stroke width, round corners, and wide (partially or entirely) enclosed areas are important elements (parts) for gaining legibility.
\end{itemize}
%==== ここまで大丈夫と思う Seiichi Feb 12
%%%%%%%%%%%%%%%%%%%%%%%%%%%%%%%%%%%%%%%%%%%%%%%%%%%%%%%%%%%
\section{Related Work}\label{sec:related}
%%%%%%%%%%%%%%%%%%%%%%%%%%%%%%%%%%%%%%%%%%%%%%%%%%%%%%%%%%%
\subsection{Font shape and impression}
As noted in Section~\ref{sec:intro}, the relationship between font shape and impression has been a topic in psychology research since the 1920's~\cite{davis1933determinants,poffenberger1923study}.  In those studies, a limited number of people provide their impression about fonts. A similar subjective analysis approach is still used even in the recent psychological studies~\cite{azadi2018multi,brumberger2003rhetoric,doyle2006dressed,grohmann2013using,henderson2004impression,mackiewicz2007audience,o2014exploratory,velasco2015taste}. For an example of the recent trials, Shaikh and Chaparro~\cite{shaikl} measured the impression of 40 fonts (10 for each of serif, sans-serif, script/handwriting, and display) by collecting the answers from 379 subjects for 16 semantic differential scales (SDS), where each SDS is a pair of antonyms, such as calm-exciting, old-young, and soft-hard. Their results clearly show the dependency of the impressions on font shape variations but do not detail what shapes are really relevant to raise an impression.\par
Computer science research has tried to realize a large-scale collection of font-impression pairs. O'Donovan et al.~\cite{o2014exploratory} use crowd-sourcing service for evaluating the grades of 37 attributes ($\sim$ impressions), such as {\it friendly}, of 200 different fonts. 
Based on this dataset, Wang et al.~\cite{Wang2020} realize a font generator called Attribute2Font
and  Choi et al.~\cite{choi2018fontmatcher} realize a font recommendation system a called FontMatcher.
Shinahara et al.~\cite{shinahara2019serif} use about 200,000 book cover images to understand the relationship between the genre and each book's title font. 
\par
More recently, Chen et al.~\cite{Chen2019large} realize a dataset that contains 18,815 fonts and 1,824 impression words, which are collected from {\tt MyFonts.com} with a cleansing process by crowd-sourcing. Since our experiment uses this dataset, we will detail it in Section~\ref{sec:dataset}. Note that the primary purpose of ~\cite{Chen2019large} is font image retrieval by an impression query and not analyze the relationship between shape and impression.
\par
Disentanglement is a technique to decompose the sample into several factors and has been applied to font images~\cite{cha2020dmfont,Liu2018,Srivatsan2019} to decompose each font image to style information and global structure information (showing the shape of each letter `A'). Although their results are promising and useful for few-shot font generation and font style transfer, their style feature has no apparent relationship neither font shape nor impression.\par
All of the above trials use the {\em whole character shape} to understand the relationship between shape and impression. Compared to parts, the whole character shape can only give a rough relationship between the shape and the impression. The whole character shape is a complex composition of strokes with local variations. We, instead, focus on parts and analyze their relationship to specific impressions more directly.

%-----------------------------------------------
\subsection{Local descriptors}
Parts of an image sample have been utilized via local descriptors, such as SIFT~\cite{SIFT} and SURF~\cite{SURF}, to realize generic object recognition, image retrieval, and image matching. Local descriptors are derived through two steps: the keypoint detection step and the description step. In the former step, parts with more geometric information, such as corner and intersection, is detected. In the later step, geometric information of each part is represented as a fixed-dimensional feature vector, which is the so-called local descriptor. In the context of generic object recognition, a set of $N$ local descriptors from an image is summarized as Bag-of-Visual Words (BoVW), where each feature vector is quantized into one of $Q$ representative feature vectors, called visual words. Then all the $L$ feature vectors are summarized as a histogram with $Q$-bins.  
\par
There are many attempts to combine local descriptors and CNNs as surveyed in \cite{Zheng2018}. However, (as shown in Table 5 of \cite{Zheng2018},) they still use BoVW summarization of the 
local descriptors. To the authors' best knowledge, this is the first attempt to combine the 
local descriptors (SIFT vectors) and DeepSets~\cite{deepsets} and learn the appropriate representation for the estimation task in an end-to-end manner. 

%==== ここまで大丈夫と思う Seiichi Feb 12
%%%%%%%%%%%%%%%%%%%%%%%%%%%%%%%%%%%%%%%%%%%%%%%%%%%%%%%%%%%
\section{Font-Impression Dataset\label{sec:dataset}}
%%%%%%%%%%%%%%%%%%%%%%%%%%%%%%%%%%%%%%%%%%%%%%%%%%%%%%%%%%%
As shown in Fig.~\ref{fig:shape-and-impression}, we use the font-impression dataset by Chen et al.~\cite{Chen2019large}. The dataset is comprised of 18,815 fonts collected from {\tt Myfonts.com}. From each font, we use 52 letter binary images of `A' to `z.' Their image size varies from 50 $\times$ 12 to 2,185 $\times$ 720. \par
For each font, $0\sim 184$ impression words are attached. The average number of impression words per font is 15.5.
The vocabulary size of the impression words is 1,824. Some of them are frequently attached to multiple fonts. For example, {\it decorative}, {\it display}, and {\it headline} are the most frequent words and attached to 6,387, 5,325, and 5,170 fonts, respectively. In contrast, some of them are rarely attached. The least frequent words ({\it web-design}, {\it harmony}, {\it jolly}, and other 13 words) are attached to only 10 fonts.\par
In the following experiment, we discarded minor impression words; specifically, if an impression word is attached to less than 100 fonts, it is discarded because it is a minor impression word with less worth for the analysis and insufficient to train our system. As a result, we consider $K=483$ impressions words.  The number of fonts was slightly decreased to 18,579 because 236 fonts have only minor impression words. The dataset was then divided into the train, validation, and test sets while following the same random and disjoint division of \cite{Chen2019large}. The numbers of fonts in these sets are 14,876, 1,856, and 1,847, respectively.
%==== ここまで大丈夫と思う Seiichi Feb 12

%%%%%%%%%%%%%%%%%%%%%%%%%%%%%%%%%%%%%%%%%%%%%%%%%%%%%%%%%%%
\section{Part-Based Impression Estimation with DeepSets}
%%%%%%%%%%%%%%%%%%%%%%%%%%%%%%%%%%%%%%%%%%%%%%%%%%%%%%%%%%%
\subsection{Extracting local shapes by SIFT}
From each image, we extract local descriptors using SIFT~\cite{SIFT}. Consequently,  we have a set of $L_i$ local descriptors, $\mathbf{X}^i = \{\mathbf{x}^i_1,\ldots,\mathbf{x}^i_{L_i}\}$, from the 52 images of the $i$th font, where $\mathbf{x}^i_l$ is a $D$-dimensional SIFT vector. The number of SIFT descriptors $L_i$ is different for each font. The average number of $L$ over all the 18,579 fonts is about $2,505$ (i.e., about 48 descriptors per letter on average)\footnote{Fonts whose stroke is filled with textures such as ``cross-hatching'' give a huge number of SIFT descriptors because they have many corners. They inflate the average number of $L$; in fact, the median of $L$ is $1,223$. In the later histogram-based analysis, we try to reduce the effect of such an extreme case by using the median-based aggregation instead of the average.}. 
Note that every SIFT vector is normalized as a unit vector, i.e., $\|\mathbf{x}^i_l\|=1$.\par
SIFT is well-known for its invariance against rotations and scales. This property might affect our analysis positively and negatively. A positive example of the rotation invariance is that it can deal with the horizontal and the vertical serifs as the same serif; a negative example is that it cannot distinguish the oblique and the upright strokes. Fortunately, our experimental results show that the negative effect is not large and the positive effect is much larger.   
%
%-----------------------------------------------
\subsection{Impression Estimation with DeepSets}
To understand the relationship between parts and impressions of fonts, we consider a task of 
estimating the impression from the set of local descriptors $\mathbf{X}^i$. The ground truth of the estimation result is represented as a $K$-dimensional $m^i$-hot vector $\mathbf{t}^i$, where $K=483$ is the vocabulary size of the impression words and $m^i$ is the number of impression words attached to the $i$-th font. 
\par
As shown in Fig.~\ref{fig:overall}, we solve this estimation task by DeepSets~\cite{deepsets}. DeepSets
first converts each $\mathbf{x}^i_l$ into another vector representation $\mathbf{y}^i_l$ by a neural network $g$; that is, $\mathbf{y}^i_l=g(\mathbf{x}^i_l)$. Then, a single sum vector $\tilde{\mathbf{y}}^i=\sum_l \mathbf{y}^i_l$ is fed to another neural network $f$ to have the 
$K$-dimensional impression vector $f(\tilde{\mathbf{y}}^i)$. By the permutation-free property 
of the summation operation, DeepSets can deal with the set $\mathbf{X}_i$ as its input.  In addition, it can deal with different $L$, without changing the network structure.  
\par
The networks $f$ and $g$ are trained in an end-to-end manner. The loss function is the binary cross-entropy between the output $f(\tilde{\mathbf{y}}^i)$ and the $m^i$-hot ground-truth  $\mathbf{t}^i$. We use a fixed-sized minibatch during training DeepSets just for computational efficiency. Each minibatch contains $64$ SIFT vectors which are randomly selected from $L_i$. In learning and inference phases, a mini-batch is created in the same way and thus gives the same computation time. Multi-Layer Perceptrons (MLPs) are used as the first and the second neural networks in the later experiment. More specifically, $g$: (128)-FC-R-(128)-FC-R-(128)-FC-(128) and $f$: (128)-T-FC-R-(256)-FC-R-(256)-FC-S(483), where FC, R, T, and S stand for fully-connected, ReLU, tanh, and sigmoid, respectively, and the parenthesized number is the dimension.
\par
%-----------------------------------------------
\subsection{Which Parts Determine a Specific Impression?}
\begin{figure}[t]
    \centering
    \includegraphics[width=\linewidth]{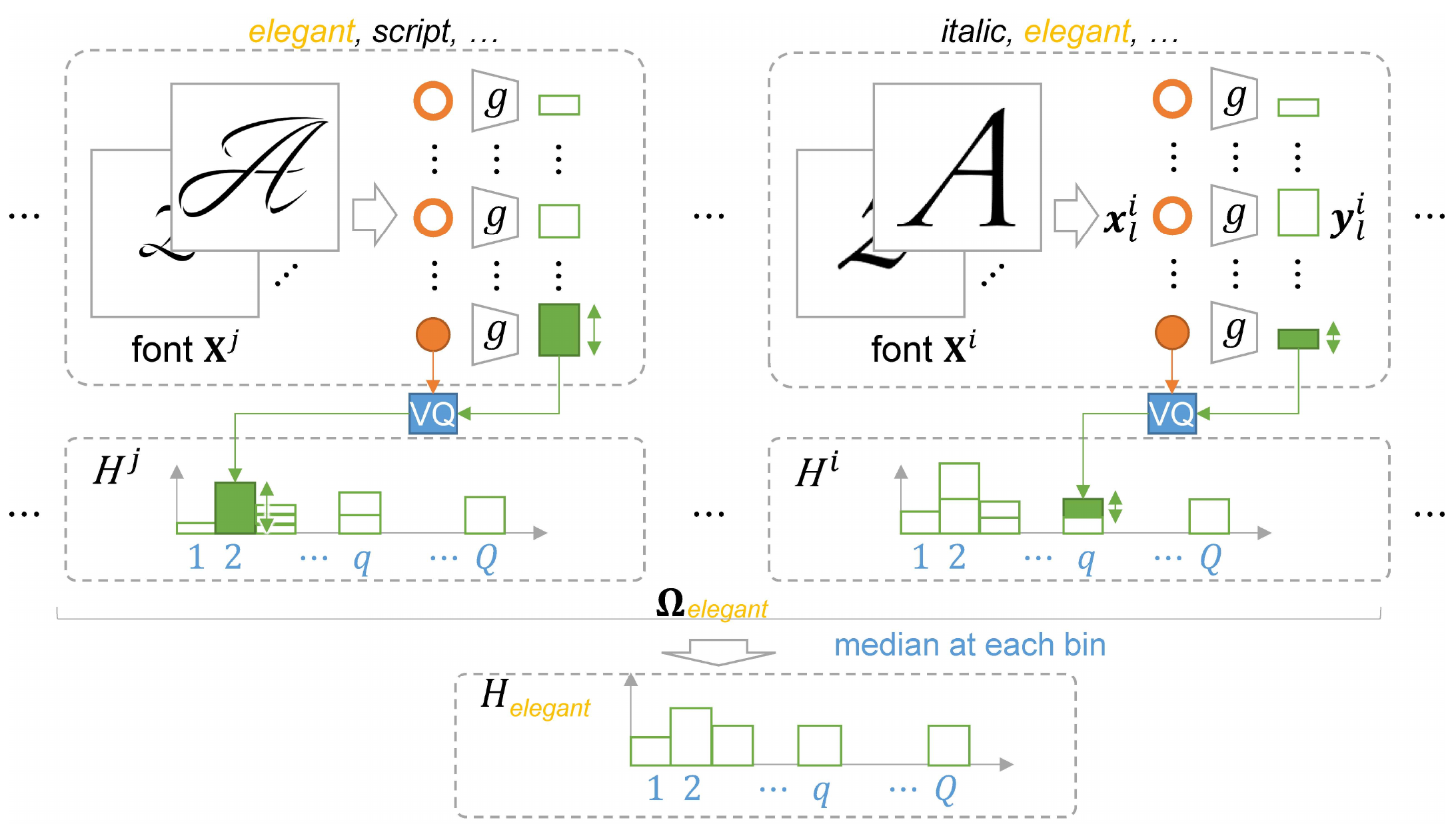}\\[-3mm]
    \caption{Understanding the important parts for a specific impression by quantization and accumulation. VQ is a quantization module where $\mathbf{x}^i_l$ is quantized to the closest one among $Q$ representative vectors. If $\mathbf{x}^i_l$ is quantized to $q$th vector, $\|\mathbf{y}^i_l\|$ is added to the $q$-th bin of the histogram $H^i$. }
    \label{fig:hist-based-analysis}
\end{figure}

An important property of DeepSets for our impression estimation task is that we can obtain the importance of individual parts by observing the intermediate outputs $\{\mathbf{y}^i_l\}$. If the $l$-th part of the $i$-th font is significant for giving 
the impression of the font, we can expect that the norm $\|\mathbf{y}^i_l\|$ will become relatively larger than the norm of unimportant parts.\footnote{Recall that the original SIFT vector is a unit vector, i.e., $\|\mathbf{x}^i_l\|=1$.} In an extreme case where the $l$-th part does not affect the impression at all, the norm $\|\mathbf{y}^i_l\|$ will become zero. \par

Fig.~\ref{fig:hist-based-analysis} illustrates its process to understand which parts are important for a specific impression $k\in [1, K]$, by using the above property.
First, for each font $i$, we create a weighted histogram $H^i$ of its SIFT feature vectors $\mathbf{X}^i$. The histogram has $Q$ bins, and each bin corresponds to a representative SIFT vector derived by k-means clustering of all SIFT vectors $\bigcup_i \mathbf{X}^i$.  
This is similar to the classical BoVW representation but different in the use of weighted votes.
If a SIFT vector $\mathbf{x}^i_l$ is quantized to the $q$-th representative vector, the norm $\|\mathbf{y}^i_l\|=\|g(\mathbf{x}^i_l)\|$ is added to the $q$-th bin. This is because different parts have different importance as indicated by the norm, which is determined automatically by DeepSets.\par
Second, for each impression $k$, we create a histogram $H_k$ by aggregating the histograms $\{H^i\}$ for $i\in \mathbf{\Omega}_k$, where $\mathbf{\Omega}_k$ denote the set of fonts annotated with the $k$-th impression.
More specifically, the histogram $H_k$ is given by $H_k = \mathrm{Med}_{i\in \mathbf{\Omega}_k} H^i$, where $\mathrm{Med}$ is the bin-wise median operation. The median-based aggregation is employed because we sometimes have a histogram $H^i$ with an impulsive peak at the $q$-th bin, and thus the effect of the single $i$-th font is overestimated in $H_k$. \par
The representative SIFT vectors with larger weighted votes in $H_k$ are evaluated as important for the $k$-th impression. This is because (1)~such SIFT vectors are evaluated as important ones with larger norms and/or (2)~such SIFT vectors frequently appear in the fonts with the impression $k$. In the later experiment, we will observe the weighted histogram $H_k$ to understand the important parts.
\par
%==== ここまで大丈夫と思う Seiichi Feb 13 morning
%%%%%%%%%%%%%%%%%%%%%%%%%%%%%%%%%%%%%%%%%%%%%%%%%%%%%%%%%%%
\section{Experimental Results}\label{sec:results}
%%%%%%%%%%%%%%%%%%%%%%%%%%%%%%%%%%%%%%%%%%%%%%%%%%%%%%%%%%%
%-----------------------------------------------
\subsection{Important parts for a specific impression}

\begin{figure}[t]
    \centering
%    \vskip 5cm
    \includegraphics[width=1.0\linewidth]{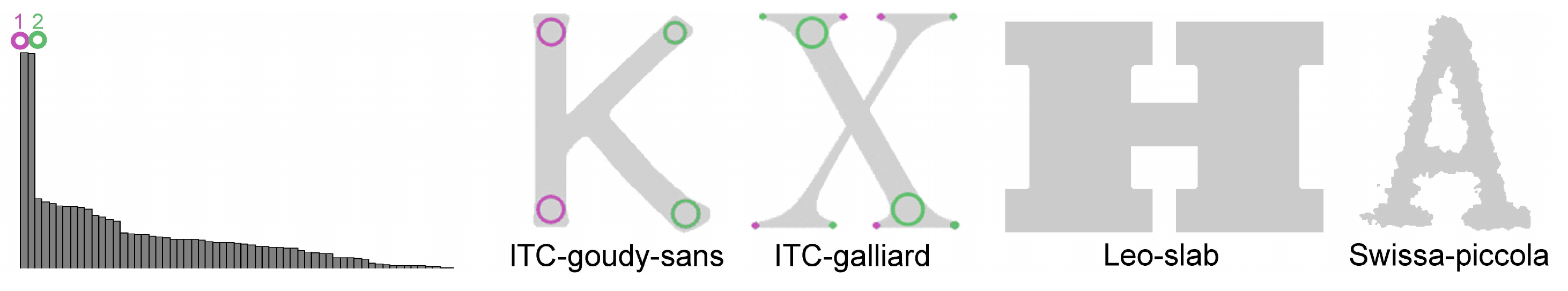}\\[-3mm]
    \caption{The average histogram $\bar{H}$. The $Q$ representative vectors are arranged in descending order of frequency. The two example images on the left show the parts that correspond to the two most frequent representative vectors.}
    \label{fig:average}
\end{figure}

Fig.~\ref{fig:average} shows the average histogram $\bar{H}=\sum_{\kappa =1}^K H_\kappa /K$, which is equivalent to the frequency of $Q$ representative vectors for all local descriptors $\bigcup_i \mathbf{X}^i$.  The representative vectors are arranged in the descending order of frequency.\par
This histogram shows that there two very frequent local shapes; the two example images in the figure show the parts that correspond to them. Those parts represented by $q=1$ and $2$ often appear near the end of straight strokes with parallel contours (i.e., strokes with a constant width), which are very common for sans-serif fonts, such as {\tt ITC-goundy-sans}, and even serif fonts, such as {\tt ITC-galliard}.  In contrast, they appear neither around extremely thick stokes even though they have a constant width, such as {\tt Leo-slab} nor the font whose stroke width varies, like {\tt Swissa-piccola}. \par

\begin{figure}[p]
    \centering
    % 認識が良好だった上位何個かのimpression classについて，
%    \vskip 5cm
    \includegraphics[width=1.0\linewidth]{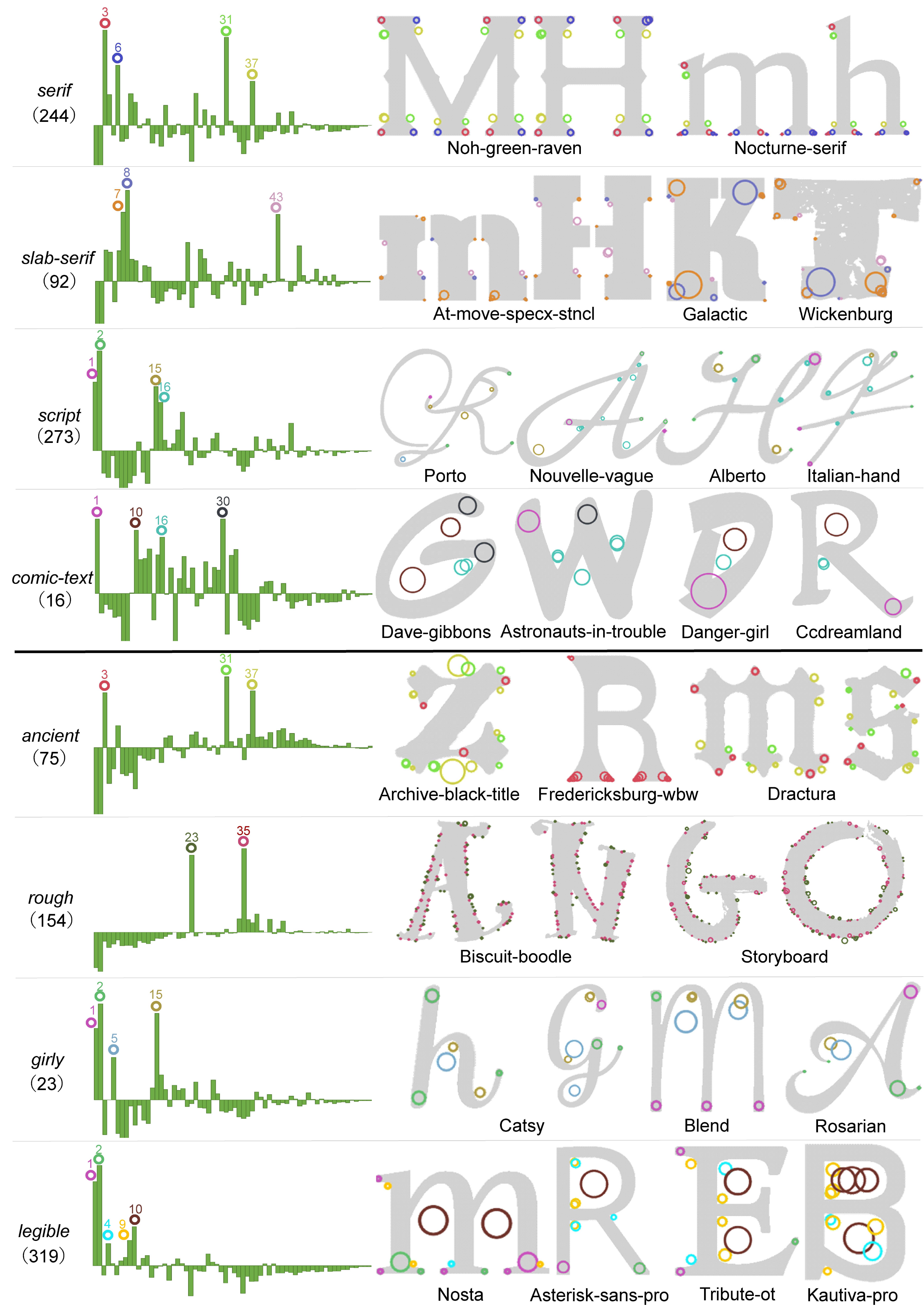}\\[-3mm]
    \caption{Which parts determine the impression of a font? --- Left: The delta-histogram $\Delta {H}_k$ and its peaks (marked by circles). Right: Several font images with the impression $k$ and the location of the important parts that correspond to the peaks. The same color is used for the peak (i.e., a representative local vector) and the corresponding parts.}
    \label{fig:main-result}
\end{figure}

Fig.~\ref{fig:main-result} visualizes the important parts for eight impressions, such as {\it serif} and {\it legible}. The parenthesized number below the impression word (such as 244 for {\it serif}) is the number of fonts with the impression, i.e., $\|\Omega_k\|$. The font name, such as {\tt Noh-green-raven}, is shown below each font image.
\par
The left-side of  Fig.~\ref{fig:main-result} shows a ``delta''-histogram $\Delta {H}_k  = H_k - \bar{H}$, which indicates the difference from the average histogram.
The positive peaks of $\Delta {H}_k$ are marked by tiny circles and their number (i.e., $q$). 
Each peak corresponds to an important representative vector for the $k$-th impression because the vectors have far larger weights than the average for $k$. 
The delta-histogram can have a negative value at the $q$-th bin when parts that give the $q$-th representative vector are less important than the average. The delta histograms $\Delta {H}_k$ suggests the different local shapes (i.e., parts) are important for different impressions because the locations of their peaks are often different. We will see later that similar impression words have similar peaks.\par
\begin{figure}[!t]
    \centering
    %\vskip 5cm
    \includegraphics[width=1\linewidth]{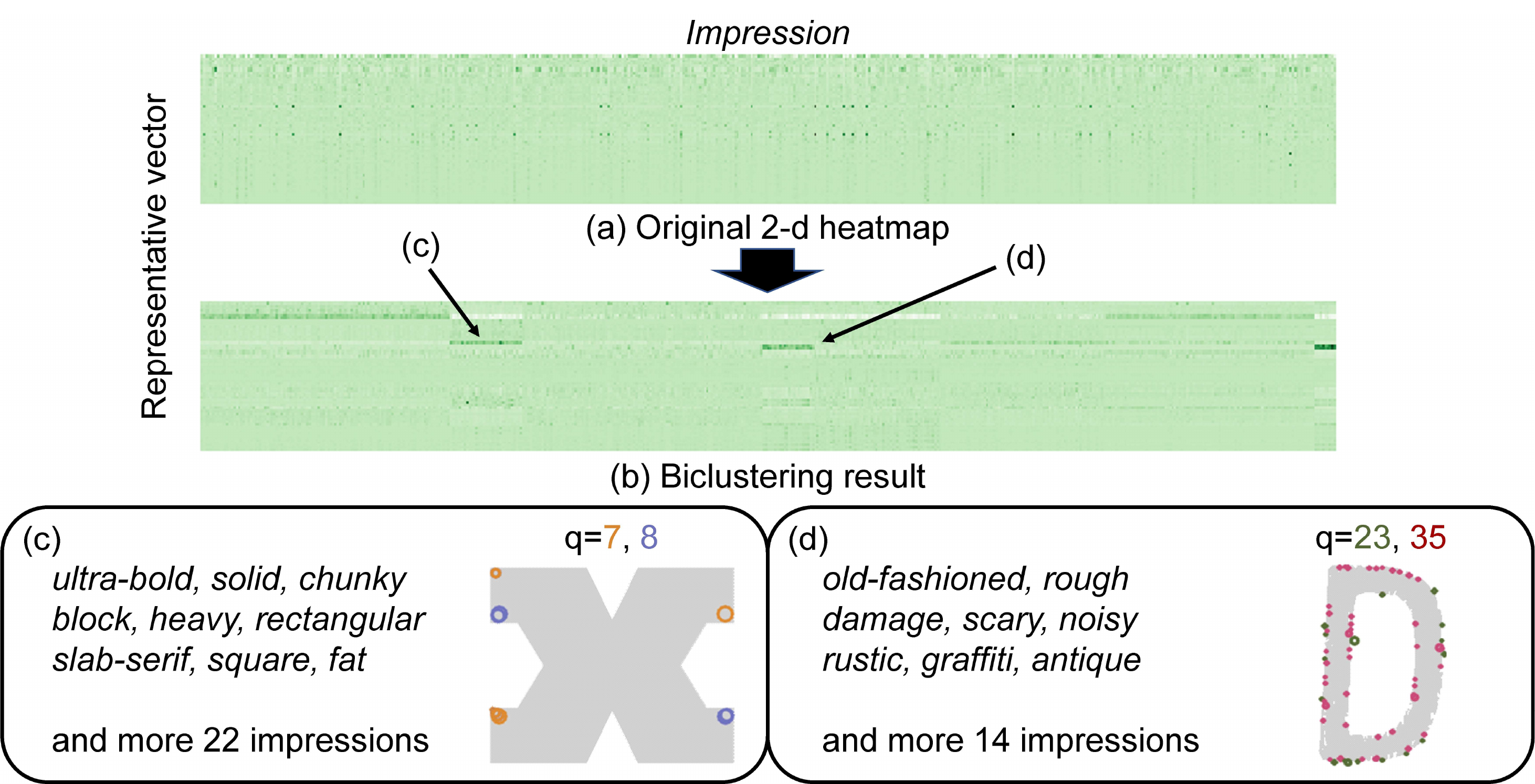}\\[-3mm]
    \caption{A biclustering result to understand the strongly-correlated pairs between  parts and impressions.}
%  \vspace{-4mm}
    \label{fig:biclustering}
%    \medskip
\end{figure}
The right-side of Fig.~\ref{fig:main-result} shows the parts corresponding to the peaks of the delta histogram on several font images --- in other words, they indicate the important parts for the impression. The top four rows are the results of less subjective impressions: 
%representative vectorの数字変更しました(←助かる！内田)
\begin{itemize}
\item {\it Serif} has peaks at $q=3,6,31,37$ that shape serifs. Precisely, the pair of $q=3$ and $6$ form the serif tips, and the other pair of $q=31$ and $37$ form the neck of the serif. It should be noted that {\it serif} has less strokes with ``parallel ending parts'' represented by $q=1$ and $2$.
\item {\it slab-serif} has extremely thick rectangular serifs and has peaks at $q=7,8$, which cannot find for {\it serif}.  
\item {\it Script} has four peaks; $q=1$ and $2$ correspond $\cup$-shaped round and parallel stroke ends. The part of $q=15$ forms an `$\ell$'-shaped curve, unique to the pen-drawn styles. $q=16$ forms an oblique stroke intersection, which can also be found in `$\ell$'-shaped stroke.
\item {\it Comic-text} has peaks at $q=10$, $16$, and $30$. The part of $q=10$ corresponds to a large partially (or entirely) enclosed area  (called ``counter'' by font-designers) formed by a circular stroke. $q=16$ is also found in {\it script} and suggests that {\it comic-text} also has an atmosphere of some handwriting styles. 
\end{itemize}\par
The four bottom rows of Fig.~\ref{fig:main-result} show the results of more subjective impressions:
\begin{itemize}
\item {\it Ancient} is has a similar $\Delta H_k$ to {\it Serif}; however, {\it ancient} does not have a peak at $q=6$, which often appears at a tip of a serif. This suggests the serif in {\it ancient} fonts are different from standard serifs.
\item {\it Rough} has two clear peaks that indicate fine jaggies of stroke contours.
\item {\it Girly} is similar to {\it script}; however,  {\it Girly} has an extra peak at $q=5$.  Comparing the parts corresponding to $q=5$ and $15$, the parts of $q=5$ show a wider curve than $15$. The parts of $q=5$ also appear at counters.
\item {\it Legible} shows its peaks at $q=1$ and $2$ and thus has parallel stroke ends. Other peaks at $q=4$ and $9$ are rounded corners. As noted at {\it comic-text}, the peak of $q=10$ shows a wider ``counter'' (i.e., enclosed area). These peaks prove the common elements in more legible fonts; strokes with constant widths, rounded corners, and wider counters.
\end{itemize}
\par
The highlight of the above observation is that it shows that we can {\em explain} the font impressions using local shapes (i.e., parts) with DeepSets and local descriptors. This fact is confirmed in the later sections where we show that
similarity between two histograms $H_k$ and $H_{k'}$ reflects the similarity between the corresponding impressions. Note that we have also confirmed that scale invariance and rotation invariance brought by SIFT greatly contributed to the analysis; the scale invariance allows us to catch the detailed and tiny local shapes. The rotation invariance allows us to identify the local shapes that are horizontally or vertically symmetrical, such as both tips of a serif.

\begin{figure}[t]
    \centering
%    \vskip 5cm
    \includegraphics[width=1\linewidth]{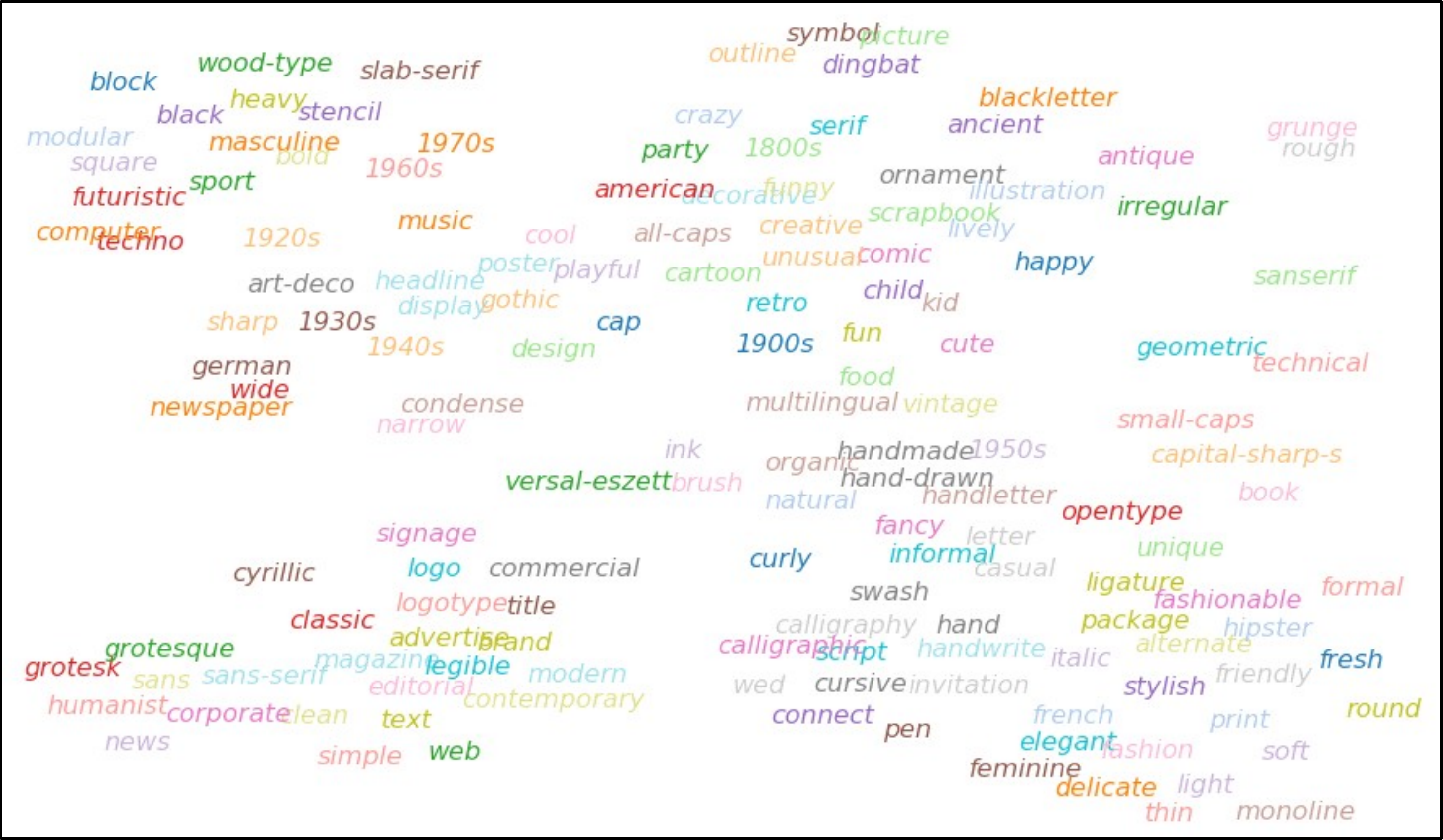}\\[-3mm]
    \caption{t-SNE visualization of the impression distributions by the similarity of the weighted histograms $\{H_k\}$. Only top-150 frequent impressions words are plotted for better visibility.  }
%    \vspace{-3mm}
    \label{fig:t-SNEmap}
\end{figure}
%-----------------------------------------------
\subsection{Parts and impressions pairs with a strong correlation}
A biclustering analysis is conducted to understand the correlation between parts and impressions more comprehensively. Fig.~\ref{fig:biclustering} shows its process and results. As shown in (a), we first prepare a matrix each of which column corresponds to $\Delta H_k$, and thus each of which row corresponds to the representative vector $q$. Then, the column  (and the row ) are re-ordered so that similar column (row) vectors come closer. Consequently, as shown in (b), the original matrix is converted as a block-like matrix, and some blocks will have larger values than the others. These blocks correspond to a strongly-correlated pair of an impression word subset and a representative vector (i.e., local shape) subset. We used the {\tt scikit-learn} implementation of spectral biclustering~\cite{biclustering}. \par

%①VW：7th, 8th（直角形状）
%印象語：['ultra-bold', 'solid', 'disco', 'chunky', 'ultra', 'russian', 'low-res', 'block', 'masculine', 'military', 'grid', 'heavy', 'rectangular', 'octagonal', 'egyptian', 'stencil', 'modular', 'bitmap', 'primitive', 'oblique', 'slab-serif', 'square', 'mechanical', 'industry', 'slab', 'wood-type', 'extend', 'dark', 'pixel', 'fat', 'black']
%②VW：23th, 35th（ガタガタのくぼみ形状）
%印象語：['brush-script', 'messy', 'weather', 'texture', 'dirty', 'distress', 'erode', 'textured', 'felt-tip','wear', 'horror', 'dry-brush', 'pencil', 'wild', 'stamp', 'old-fashioned', 'rough', 'damage', 'paint', 'ghost', 'grunge', 'noisy', 'postcard', 'graffiti', 'tag', 'rustic', 'punk', 'break', 'useful', 'antique', 'scary']
Among several highly correlating blocks in Fig.~\ref{fig:biclustering}~(b), two very prominent blocks are presented as (c) and (d). The block (c) suggests that the impressions showing solid and square atmospheres strongly correlate to the local shapes represented by $q=7$ and $8$. The block (d) suggests that the impressions showing jaggy and old-fashioned atmospheres strongly correlate to $q=23$ and $35$. This analysis result shows that our strategy using parts has a very high explainability for understanding the correlations between the part and impressions.

\begin{figure}[t]
    \centering
%    \vskip 5cm
    \includegraphics[width=1\linewidth]{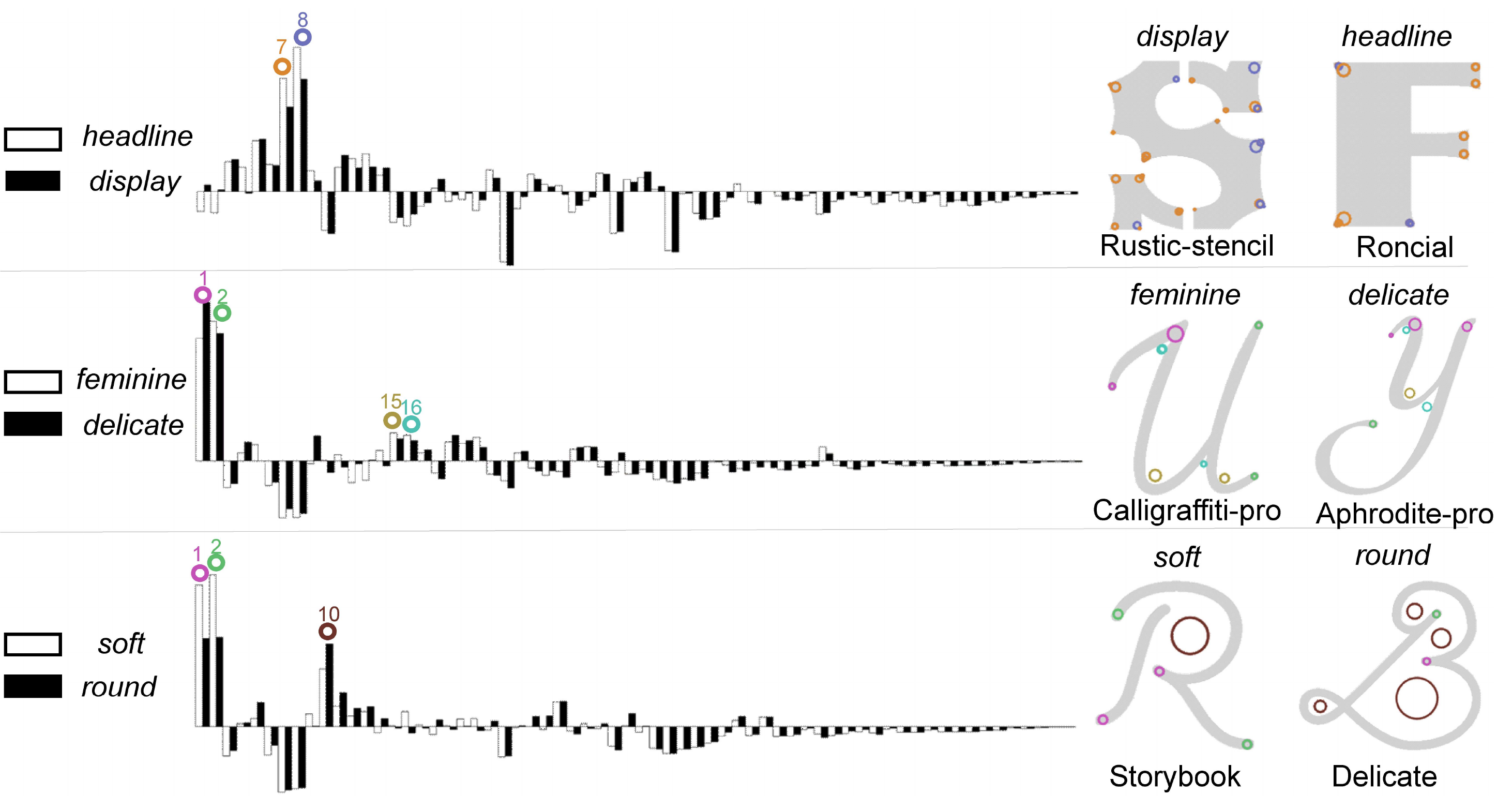}\\[-3mm]
    \caption{The delta-histograms $\Delta H_k$ for similar impression words.}
%    \vspace{-4mm}
    \label{fig:each-impression-font-shape}
\end{figure}
%-----------------------------------------------
\subsection{Similarity among impression words by parts}
Fig.~\ref{fig:t-SNEmap} shows the t-SNE visualization of the distributions of the weighted histograms $\{H_k\}$ (or, equivalently, $\{\Delta H_k\}$).
For better visibility and higher reliability, the top 150 frequent impression words are selected and plotted.
This plot clearly shows that similar impression words are often close to each other. For example, {\it calligraphic}, {\it script} and {\it cursive} are neighboring because they imply a font with curvy strokes.
({\it bold}, {\it heavy}) and ({\it soft}, {\it round}) are also neighboring. In fact, by looking only less subjective words (i.e., words of typography), this distribution are roughly divided into four regions: bold sans-serif fonts (top-left), regular sans-serif fonts (bottom-left), script and handwriting font (bottom-right), and serif fonts (top-right).\par
More subjective impression words also form a cluster. For example, {\it delicate}, {\it elegant}, {\it french}, and {\it feminine} are close to each other, and belong to the ``script and handwriting font'' region. {\it Legible}, {\it modern}, and {\it brand} are also very close
and belong to the ``regular sans-serif'' region. {\it Cute}, {\it kid}, and {\it fun} belong to an intermediate region. Note that we can also observe the history of font trends by watching the transition from {\it 1800s} to {\it 1970s}.  According to those observations, Fig.~\ref{fig:t-SNEmap} is confirmed as the first large evidence-based proof that local shapes and impressions clearly correlate. \par
Fig.~\ref{fig:each-impression-font-shape} shows $H_k$s of a pair of similar words for confirming the above observation. Each pair has very similar histograms by sharing the same peak locations.

\begin{table}[t]
\caption{The top 20 impression words in average precision (AP). For each impression word, 5 font examples are shown.}
\label{table:good-word-list}
%\vspace{-3mm}
\begin{minipage}[t]{0.49\linewidth}
    \begin{tabular}{c|c|c|c}
        Rank & Impression & AP(\%) & font examples \\ \hline\hline
        1 & {\it sans-serif} & 68.66 & 
        \begin{minipage}{20mm}
        \centering
%        \vspace{0.5mm}
    	\includegraphics[width=20mm]{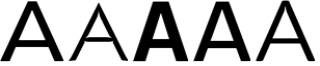}
        \end{minipage} \\ \hline
    	2 &  {\it handwrite} & 68.55 &
        \begin{minipage}{20mm}
        \centering
%        \vspace{0.5mm}
    	\includegraphics[width=20mm]{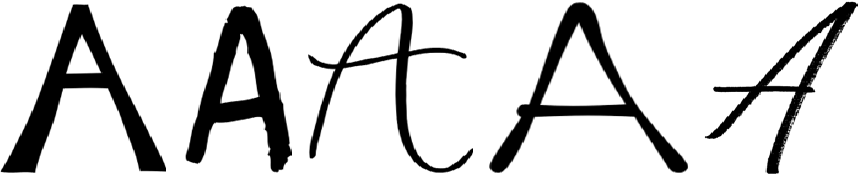}
        \end{minipage} \\ \hline
        3 &  {\it script} & 67.80 &
        \begin{minipage}{20mm}
        \centering
%        \vspace{0.5mm}
    	\includegraphics[width=20mm]{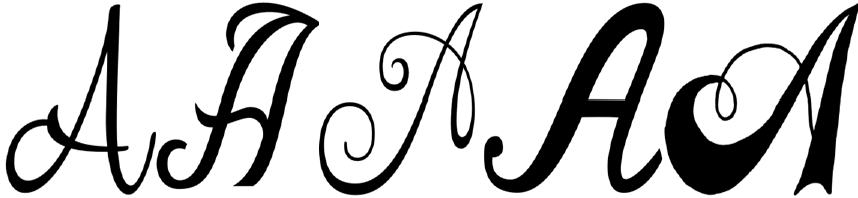}
        \end{minipage} \\ \hline
        4 &  {\it black-letter} & 66.71 &
        \begin{minipage}{20mm}
        \centering
%        \vspace{0.5mm}
    	\includegraphics[width=20mm]{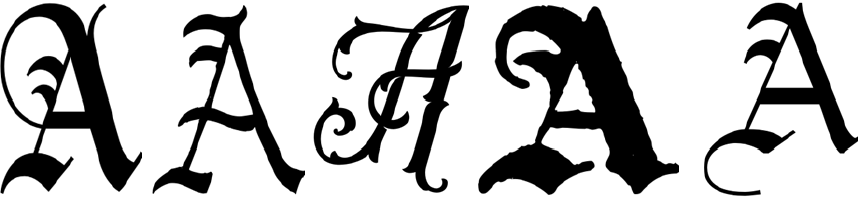}
        \end{minipage} \\ \hline
        5 &  {\it slab-serif} & 63.95 &
        \begin{minipage}{20mm}
        \centering
%        \vspace{0.5mm}
    	\includegraphics[width=20mm]{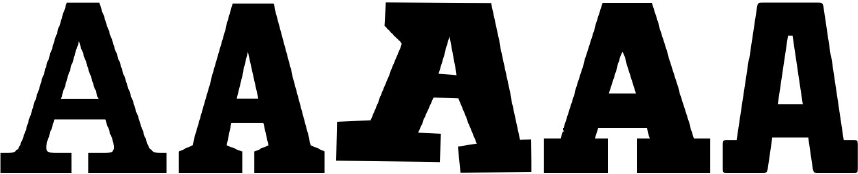}
        \end{minipage} \\ \hline
        6 &  {\it serif} & 60.00 &
        \begin{minipage}{20mm}
        \centering
%        \vspace{0.5mm}
    	\includegraphics[width=20mm]{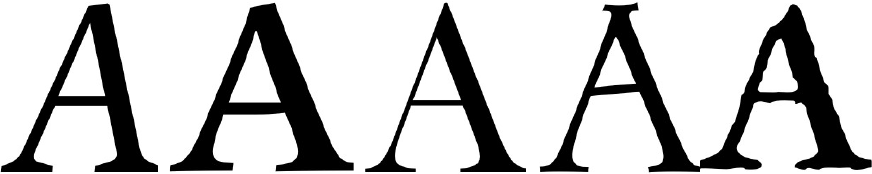}
        \end{minipage} \\ \hline
        7 & {\it  text} & 57.30 &
        \begin{minipage}{20mm}
        \centering
%        \vspace{0.5mm}
    	\includegraphics[width=20mm]{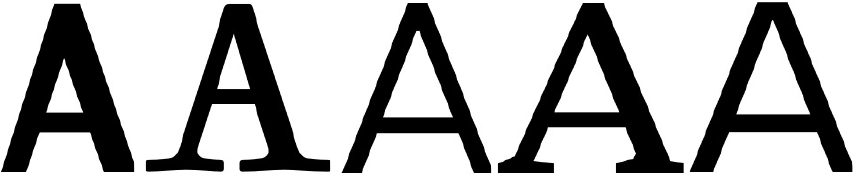}
        \end{minipage} \\ \hline
        8 &  {\it round} & 55.62 &
        \begin{minipage}{20mm}
        \centering
%        \vspace{0.5mm}
    	\includegraphics[width=20mm]{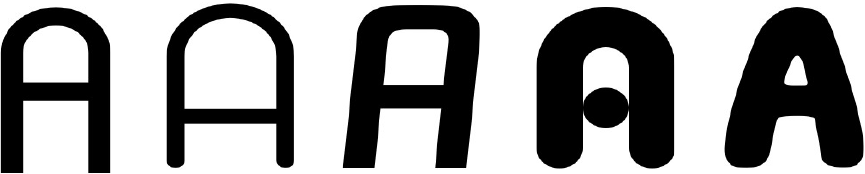}
        \end{minipage} \\ \hline
        9 &  {\it grunge} & 54.16 &
        \begin{minipage}{20mm}
        \centering
%        \vspace{0.5mm}
    	\includegraphics[width=20mm]{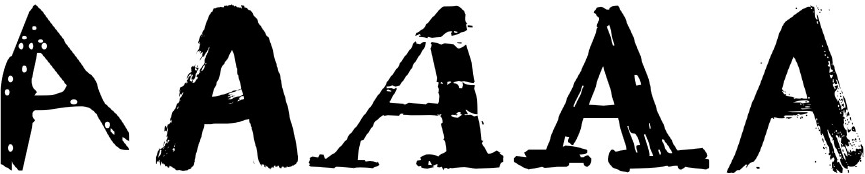}
        \end{minipage} \\ \hline
        10 &  {\it rough} & 52.82 &
        \begin{minipage}{20mm}
        \centering
%        \vspace{0.5mm}
    	\includegraphics[width=20mm]{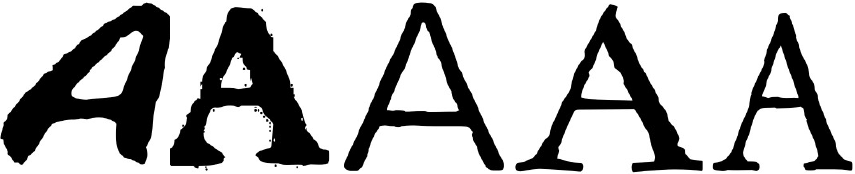}
        \end{minipage} \\ \hline
    \end{tabular}
  \end{minipage}
\hfill
\begin{minipage}[t]{0.49\linewidth}
    \begin{tabular}{c|c|c|c}
        Rank & Impression & AP(\%) & font examples \\ \hline\hline
        11 &  {\it bitmap} & 50.70 &
        \begin{minipage}{20mm}
        \centering
%        \vspace{0.5mm}
    	\includegraphics[width=20mm]{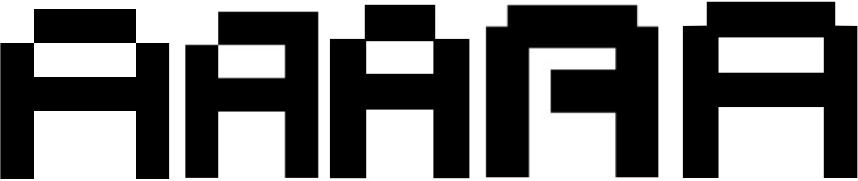}
        \end{minipage} \\ \hline
    	12 &  {\it stencil} & 49.90 &
        \begin{minipage}{20mm}
        \centering
%        \vspace{0.5mm}
    	\includegraphics[width=20mm]{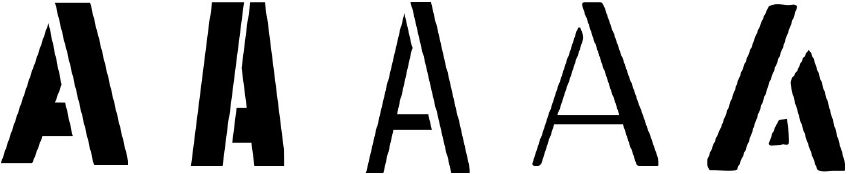}
        \end{minipage} \\ \hline
        13 &  {\it comic-text} & 49.10 &
        \begin{minipage}{20mm}
        \centering
%        \vspace{0.5mm}
    	\includegraphics[width=20mm]{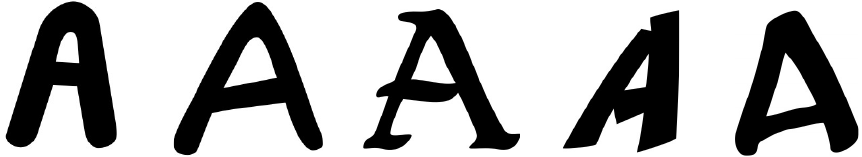}
        \end{minipage} \\ \hline
        14 &  {\it elegant} & 48.91 &
        \begin{minipage}{20mm}
        \centering
%        \vspace{0.5mm}
    	\includegraphics[width=20mm]{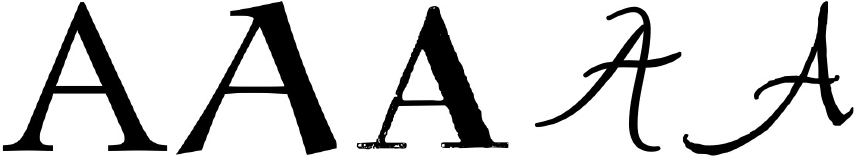}
        \end{minipage} \\ \hline
        15 &  {\it magazine} & 48.79 &
        \begin{minipage}{20mm}
        \centering
%        \vspace{0.5mm}
    	\includegraphics[width=20mm]{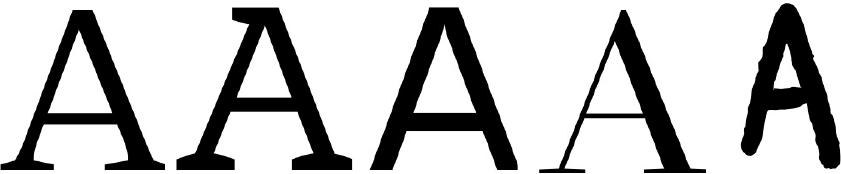}
        \end{minipage} \\ \hline
        16 &  {\it decorative} & 48.78 &
        \begin{minipage}{20mm}
        \centering
%        \vspace{0.5mm}
    	\includegraphics[width=20mm]{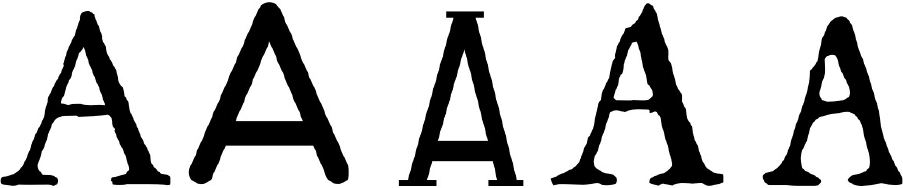}
        \end{minipage} \\ \hline
        17 &  {\it legible} & 48.78 &
        \begin{minipage}{20mm}
        \centering
%        \vspace{0.5mm}
    	\includegraphics[width=20mm]{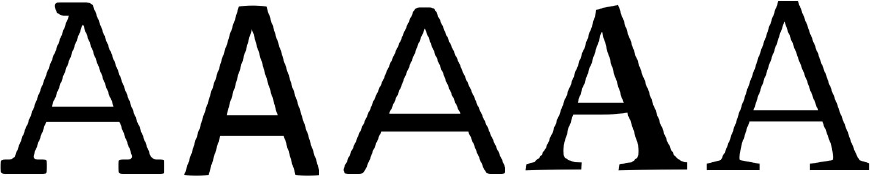}
        \end{minipage} \\ \hline
        18 &  {\it headline} & 47.68 &
        \begin{minipage}{20mm}
        \centering
%        \vspace{0.5mm}
    	\includegraphics[width=20mm]{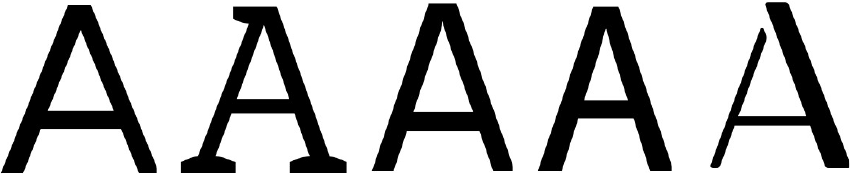}
        \end{minipage} \\ \hline
        19 &  {\it didone} & 46.80 &
        \begin{minipage}{20mm}
        \centering
%        \vspace{0.5mm}
    	\includegraphics[width=20mm]{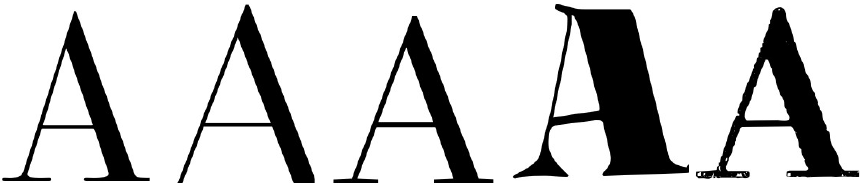}
        \end{minipage} \\ \hline
        20 &  {\it brush} & 45.54 &
        \begin{minipage}{20mm}
        \centering
%        \vspace{0.5mm}
    	\includegraphics[width=20mm]{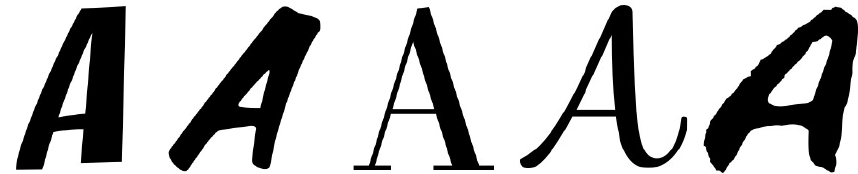}
        \end{minipage} \\ \hline
    \end{tabular}
\end{minipage}
%\end{table}
%\begin{table}[t]
\bigskip
\bigskip
\bigskip
\caption{The bottom 20 impression words in average precision (AP).}
\label{table:bad-word-list}
%\vspace{-3mm}
\begin{minipage}[t]{0.49\linewidth}
    \begin{tabular}{c|c|c|c}
        Rank & Impression & AP(\%)  & font examples \\ \hline\hline
        464 &  {\it 2000s} & 2.04 &
        \begin{minipage}{20mm}
        \centering
%        \vspace{0.5mm}
    	\includegraphics[width=20mm]{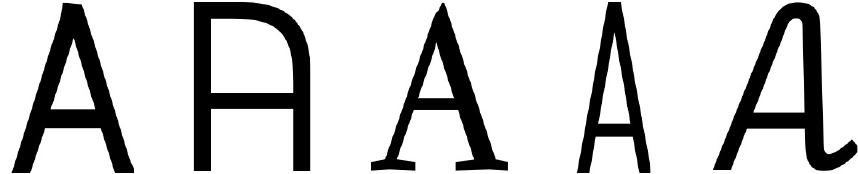}
        \end{minipage} \\ \hline
    	465 &  {\it heart} & 2.03 &
        \begin{minipage}{20mm}
        \centering
%        \vspace{1mm}
    	\includegraphics[width=20mm]{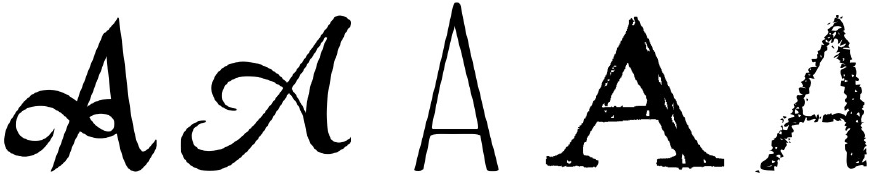}
        \end{minipage} \\ \hline
        466 &  {\it random} & 2.01 &
        \begin{minipage}{20mm}
        \centering
%        \vspace{1mm}
    	\includegraphics[width=20mm]{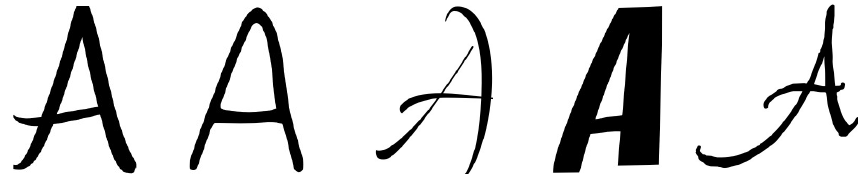}
        \end{minipage} \\ \hline
        467 &  {\it curve} & 1.95 &
        \begin{minipage}{20mm}
        \centering
%        \vspace{1mm}
    	\includegraphics[width=20mm]{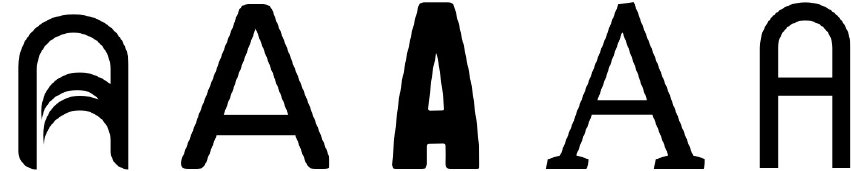}
        \end{minipage} \\ \hline
        468 &  {\it fantasy} & 1.92 &
        \begin{minipage}{20mm}
        \centering
%        \vspace{1mm}
    	\includegraphics[width=20mm]{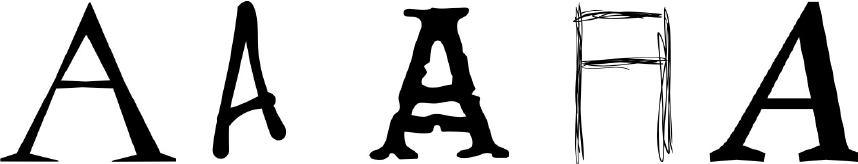}
        \end{minipage} \\ \hline
        469 &  {\it arrow} & 1.91 &
        \begin{minipage}{20mm}
        \centering
%        \vspace{1mm}
    	\includegraphics[width=20mm]{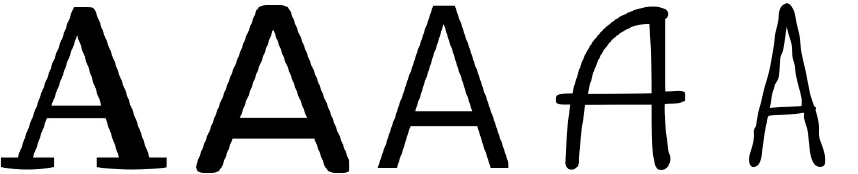}
        \end{minipage} \\ \hline
        470 &  {\it girl} & 1.90 &
        \begin{minipage}{20mm}
        \centering
%        \vspace{1mm}
    	\includegraphics[width=20mm]{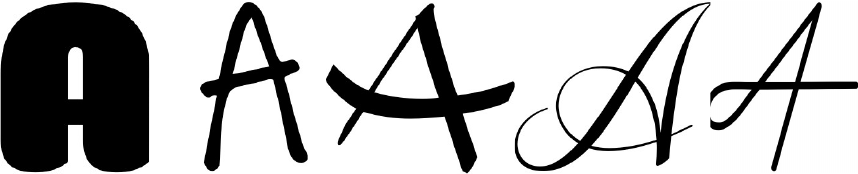}
        \end{minipage} \\ \hline
        471 &  {\it new} & 1.86 &
        \begin{minipage}{20mm}
        \centering
%        \vspace{1mm}
    	\includegraphics[width=20mm]{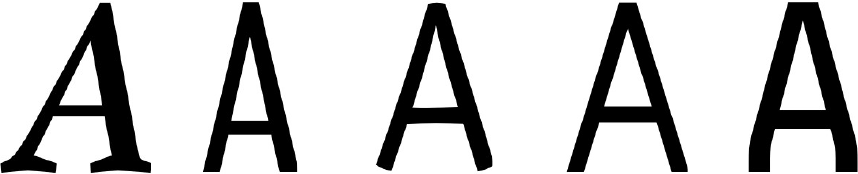}
        \end{minipage} \\ \hline
        472 &  {\it circle} & 1.81 &
        \begin{minipage}{20mm}
        \centering
%        \vspace{1mm}
    	\includegraphics[width=20mm]{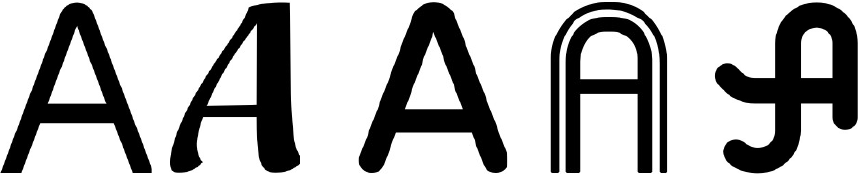}
        \end{minipage} \\ \hline
        473 &  {\it package} & 1.76 &
        \begin{minipage}{20mm}
        \centering
%        \vspace{1mm}
    	\includegraphics[width=20mm]{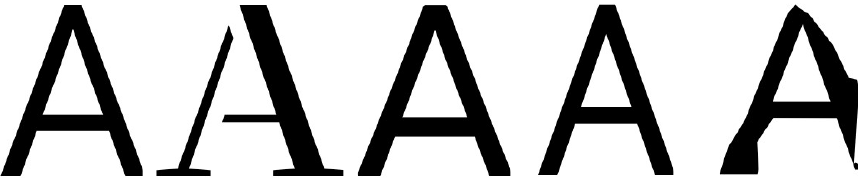}
        \end{minipage} \\ \hline
    \end{tabular}
  \end{minipage}
  \hfill
\begin{minipage}[t]{0.49\linewidth}
    \begin{tabular}{c|c|c|c}
        Rank & Impression & AP(\%)  & font examples \\ \hline\hline
        474 &  {\it graphic} & 1.71 &
        \begin{minipage}{20mm}
        \centering
%        \vspace{0.5mm}
    	\includegraphics[width=20mm]{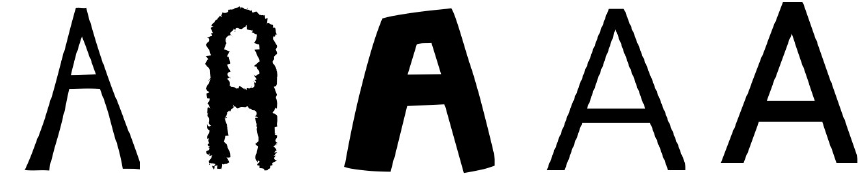}
        \end{minipage} \\ \hline
    	475 &  {\it sassy} & 1.64 &
        \begin{minipage}{20mm}
        \centering
%        \vspace{1mm}
    	\includegraphics[width=20mm]{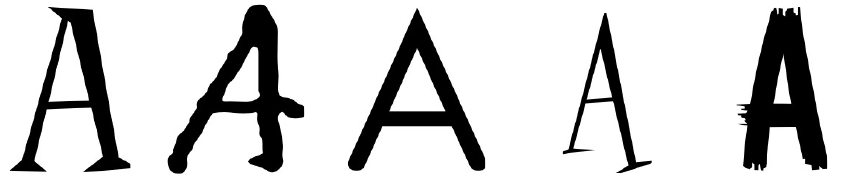}
        \end{minipage} \\ \hline
        476 &  {\it distinctive} & 1.58 &
        \begin{minipage}{20mm}
        \centering
%        \vspace{1mm}
    	\includegraphics[width=20mm]{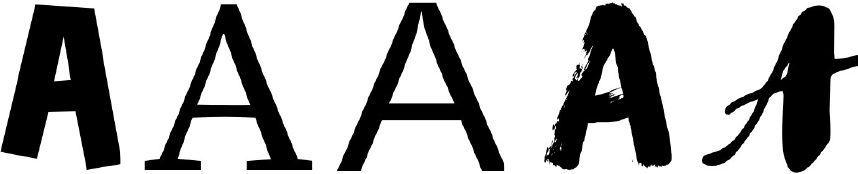}
        \end{minipage} \\ \hline
        477 &  {\it style} & 1.57 &
        \begin{minipage}{20mm}
        \centering
%        \vspace{1mm}
    	\includegraphics[width=20mm]{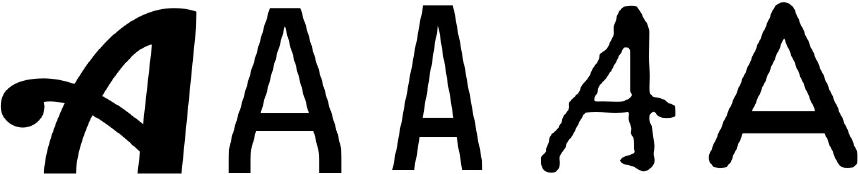}
        \end{minipage} \\ \hline
        478 &  {\it magic} & 1.45 &
        \begin{minipage}{20mm}
        \centering
%        \vspace{1mm}
    	\includegraphics[width=20mm]{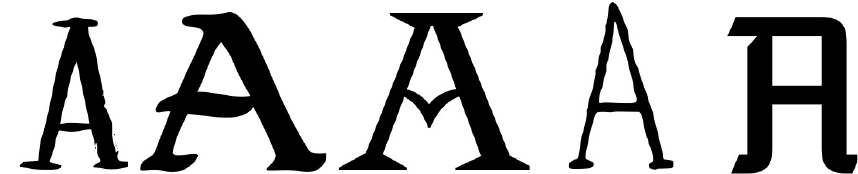}
        \end{minipage} \\ \hline
        479 &  {\it chic} & 1.44 &
        \begin{minipage}{20mm}
        \centering
%        \vspace{1mm}
    	\includegraphics[width=20mm]{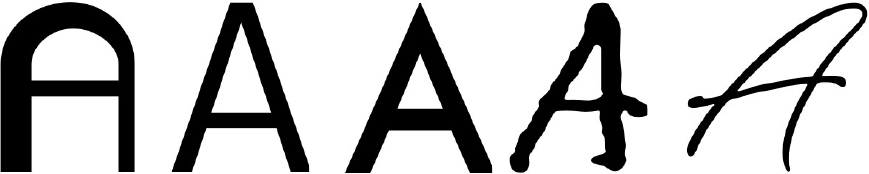}
        \end{minipage} \\ \hline
        480 &  {\it revival} & 1.34 &
        \begin{minipage}{20mm}
        \centering
%        \vspace{1mm}
    	\includegraphics[width=20mm]{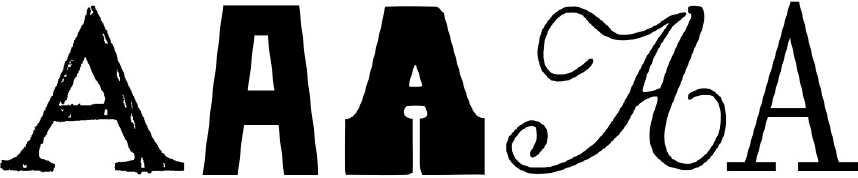}
        \end{minipage} \\ \hline
        481 &  {\it oblique} & 1.18 &
        \begin{minipage}{20mm}
        \centering
%        \vspace{1mm}
    	\includegraphics[width=20mm]{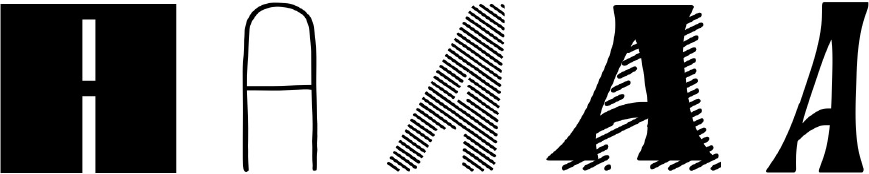}
        \end{minipage} \\ \hline
        482 &  {\it thick} & 1.13 &
        \begin{minipage}{20mm}
        \centering
%        \vspace{1mm}
    	\includegraphics[width=20mm]{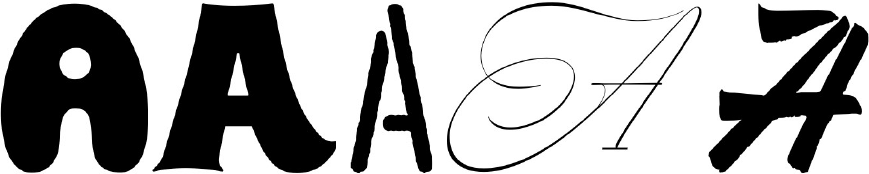}
        \end{minipage} \\ \hline
        483 &  {\it travel} & 0.64 &
        \begin{minipage}{20mm}
        \centering
%        \vspace{1mm}
    	\includegraphics[width=20mm]{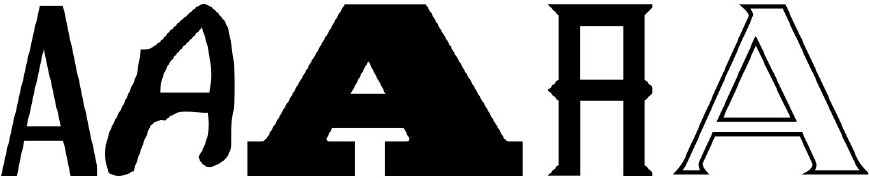}
        \end{minipage} \\ \hline
    \end{tabular}
\end{minipage}
%\vspace{-5mm}
\end{table}

%-----------------------------------------------
\subsection{Evaluating impression stability by estimation accuracy}
As we see in Fig.~\ref{fig:overall}, DeepSets is trained to estimate the impression words for a given set of SIFT vectors. By using the estimation accuracy on the {\em test set}, we can understand the stability of the impressions. When an impression word $k$ is estimated accurately,  the impression $k$ is stably correlating to the parts of the fonts with the impression $k$. 
For the quantitative evaluation of the estimation accuracy, we use average precision (AP) for each impression $k$. AP for $k$ is given as $(\sum_{h=1}^{|\mathbf{\Omega}_k|}h/r_h)/|\mathbf{\Omega}_k|$, where $r_h$ is the rank of the font $h \in \mathbf{\Omega}_k$ in the list of the likelihoods of the impression of $k$; AP for $k$ becomes larger when the fonts with the impression $k$ gets higher likelihoods of $k$ by DeepSets.\par
Table~\ref{table:good-word-list} shows the impression words with the 20 highest AP values and five font images with the impressions. This table indicates that the impression words for describing font shapes less subjectively are more stable. Especially, top-6 words (from {\it sans-serif} to {\it serif}) are technical terms for font designs and corresponding to specific font shapes. This fact proves that our analysis with DeepSets is reasonable.\par
Table~\ref{table:good-word-list} also shows that more subjective impressions can have a stable correspondence with parts. For example, {\it grunge} (9th), {\it rough} (10th), {\it elegant} (14th)  and {\it legible} (17th) have a high AP and thus those impressions are clearly determined by the set of parts of a font image. The high AP of {\it legible} indicates that better legibility is common for many people. We also share similar elegant impressions from fonts with specific parts.
It is also shown that parts can specify certain styles. For example, {\it comic-text} (13th), {\it magazine} (15th) , and  {\it headline} (18th) are styles that are determined by parts. This means that we can imagine similar fonts that are suitable for comics and headlines. \par
Table~\ref{table:bad-word-list} shows 20 impression words with the lowest AP values. Those impression words are least stable in their local parts. In other words, it is not easy to estimate the impression from parts. Instability of several impression words in the table is intuitively understandable; for example, it is difficult to imagine some valid font shapes 
from vague impressions, such as {\it random} (466th), {\it new} (471th), {\it graphic} (474th), and {\it style} (477th). It should be noted that the words relating to some specific shapes, such as {\it curve}, {\it circle}, {\it oblique}, and {\it thick}, are also listed in this table.
This means that those shapes have large variations and are not stable, at least, in their parts. In other words, these shape-related impression words are not suitable as a query for searching fonts because the search results show too large diversities.\par
It would be interesting to see how much accuracy we can get if we use a simple CNN-based model for the impression word classification rather than using SIFT. It would tell us the expressive ability of the SIFT-based method. We conducted a experiment to recognize the impression by inputting the whole font image to a standard CNN (ResNet50\cite{resnet}). 
Since the whole image conveys more information than a set of its local parts (more specifically, the whole image can use the absolute and relative location of its local parts), the use of the whole image gave a better mean Average Precision(mAP) as expected. More precisely, the standard CNN approach with the whole image achieved about 5.95-point higher mAP than our part-based method. 
However, it should be emphasized that the use of the whole font image cannot answer our main question – where the impression comes from. In other words, our SIFT-based approach can explain the relevant local parts very explicitly with the cost of 5.95-point degradation.
%We believe our explainability is far more explicit and accurate than the recent XAI techniques, such as GradCAM.
\par

%%%%%%%%%%%%%%%%%%%%%%%%%%%%%%%%%%%%%%%%%%%%%%%%%%%%%%%%%%%
\section{Conclusion}\label{sec:conclusion}
%%%%%%%%%%%%%%%%%%%%%%%%%%%%%%%%%%%%%%%%%%%%%%%%%%%%%%%%%%%
This paper analyzed the correlation between the parts (local shapes) and the impressions of fonts by newly combining SIFT and DeepSets. SIFT is used to extract an arbitrary number of essential parts from a particular font. 
DeepSets are used to summarize the parts into a single impression vector 
with appropriate weights. The weights are used for an index for the importance of the part for an impression. Various correlation analyses with 18,579 fonts and 483 impression words from the dataset~\cite{Chen2019large} prove that our part-based analysis strategy gives clear explanations about the correlations, even though it still utilizes representation learning in DeepSets for dealing with the nonlinear correlations. Our results will be useful to generate new fonts with specific impressions and solve open problems, such as what legibility is and what elegance is in fonts.
%%%%%%%%%%%%%%%%%%%%%%%%%%%%%%%%%%%%%%%%%%%%%%%%%%%%%%%%%%%
\section*{Acknowledgment}
%\vspace{-3mm}
%%%%%%%%%%%%%%%%%%%%%%%%%%%%%%%%%%%%%%%%%%%%%%%%%%%%%%%%%%%
This work was supported by JSPS KAKENHI Grant Number JP17H06100.
%

\bibliographystyle{splncs04}
\bibliography{icdar}




% LNCS Homepage, \url{http://www.springer.com/lncs}. Last accessed 4
% \end{thebibliography}
\end{document}